\documentclass{article}

\usepackage{mystyle}

\usepackage[utf8]{inputenc} 
\usepackage[T1]{fontenc}    
\usepackage{hyperref}       
\usepackage{url}            
\usepackage{booktabs}       
\usepackage{amsfonts}       
\usepackage{nicefrac}       
\usepackage{microtype}      
\usepackage{amssymb}        
\usepackage{amsmath}
\usepackage{cleveref}       
\usepackage{lipsum}         
\usepackage{graphicx}
\usepackage[numbers,sort&compress]{natbib}
\usepackage{doi}
\usepackage{multirow}

\usepackage{enumitem}
\setlist[itemize]{
    topsep=3pt,      
    itemsep=2pt,     
    parsep=0pt,      
    leftmargin=15pt  
}
\setlist[enumerate]{
    topsep=3pt,
    itemsep=2pt,
    parsep=0pt,
    leftmargin=15pt
}

\usepackage[font=small,skip=4pt]{caption} 
\title{MatQnA: A Benchmark Dataset for Multi-modal Large Language Models in Materials Characterization and Analysis}

\date{}

\newif\ifuniqueAffiliation
\uniqueAffiliationtrue

\ifuniqueAffiliation 
\author{ 
    Yonghao Weng\thanks{Scientific Compass(www.shiyanjia.com) is a leading one-stop comprehensive scientific research service platform in China, providing professional technical support and solutions (covering material characterization, chemical analysis, biological testing, etc.) for researchers, universities and enterprises.} \\
    Department of Materials Engineering\\
    Zhejiang University\\
    Hangzhou, Zhejiang Province 310000 \\
    \texttt{22460356@zju.edu.cn} \\
    \And
    Linwu Zhu \\
    Department of Data Intelligence\\
    Shiyanjia Lab of Scientiﬁc Compass\\
    Hangzhou, Zhejiang Province 310000 \\
    \texttt{zhulinwu@shiyanjia.com} \\
    \And
    Liqiang Gao \\
    Department of Data Intelligence\\
    Shiyanjia Lab of Scientiﬁc Compass\\
    Hangzhou, Zhejiang Province 310000 \\
    \texttt{gaoliqiang@shiyanjia.com} \\
    \And
    Jian Huang \\
    Department of Data Intelligence\\
    Shiyanjia Lab of Scientiﬁc Compass\\
    Hangzhou, Zhejiang Province 310000 \\
    \texttt{huangjian@shiyanjia.com} \\
}
\else
\usepackage{authblk}

\author[1]{
    Yonghao Weng\thanks{\texttt{22460356@zju.edu.cn}}%
}
\author[1,2]{
    Liqiang Gao\thanks{\texttt{gaoliqiang@shiyanjia.com}}%
}
\affil[1]{Department of Materials Science and Engineering, Zhejiang University, Hangzhou, Zhejiang Province 310000}
\affil[2]{Department of Data Intelligence, Shiyanjia Lab of Scientiﬁc Compass, Hangzhou, Zhejiang Province 310000}
\fi

\renewcommand{\headeright}{}
\renewcommand{\undertitle}{}
\renewcommand{\shorttitle}{}

\begin{document}
\maketitle

\begin{abstract}
    Recently, large language models (LLMs) have achieved remarkable breakthroughs in general domains such as programming and writing, and have demonstrated strong potential in various scientific research scenarios. However, the capabilities of AI models in the highly specialized field of materials characterization and analysis have not yet been systematically or sufficiently validated. To address this gap, we present \textbf{MatQnA}, the first multi-modal benchmark dataset specifically designed for material characterization techniques. MatQnA includes ten mainstream characterization methods, such as X-ray Photoelectron Spectroscopy (XPS), X-ray Diffraction (XRD), Scanning Electron Microscopy (SEM), Transmission Electron Microscopy (TEM), etc. We employ a hybrid approach combining LLMs with human-in-the-loop validation to construct high-quality question-answer pairs, integrating both multiple-choice and subjective questions. Our preliminary evaluation results show that the most advanced multi-modal AI models (e.g., GPT-4.1, Claude 4, Gemini 2.5, and Doubao Vision Pro 32K) have already achieved nearly 90\% accuracy on objective questions in materials data interpretation and analysis tasks, demonstrating strong potential for applications in materials characterization and analysis. The MatQnA dataset is publicly available at \url{https://huggingface.co/datasets/richardhzgg/matQnA}.

\end{abstract}


\section{INTRODUCTION}
Large Language Models (LLMs) are reshaping the technological landscape of Natural Language Processing (NLP). By analyzing vast amounts of textual data, these models demonstrate strong capabilities in pattern recognition, outcome prediction, and semantic generation. The potential of LLMs extends far beyond traditional NLP tasks, with their influence now reaching into specialized fields such as medical diagnostics ~\cite{singhal2023large} and financial analysis ~\cite{bloomberggpt2023}. This underscores the cross-domain adaptability of general-purpose artificial intelligence. In line with this trend, the field of materials science is also encountering transformative opportunities brought about by LLMs. Recent studies suggest that LLMs hold great promise in key areas including materials discovery, property prediction, experimental design optimization, multimodal data integration, and knowledge graph construction ~\cite{liu2025mattools}.

The effectiveness of LLMs in specialized domains fundamentally depends on their grasp of domain-specific knowledge. To systematically evaluate and enhance this domain expertise, the establishment of scientific evaluation systems is essential. In this context, the importance of domain-specific benchmark datasets has become increasingly evident. Such datasets provide a standardized framework for quantifying LLMs' proficiency in specific fields. For instance, in NLP, datasets like SWAG ~\cite{zellers2018swag} and SocialIQa ~\cite{sap2019socialiqa} are designed to evaluate situational reasoning and social intelligence, respectively. Similarly, in medicine and finance, benchmarks such as MedQA ~\cite{Jin2021MedQA} and BloombergGPT ~\cite{bloomberggpt2023} serve as critical references for model optimization and assessment.

LLMs are also increasingly applied in the field of materials science, yet there is a significant lack of benchmark datasets tailored to evaluating their domain-specific capabilities in materials characterization and analysis~\cite{liu2025mattools}. This gap limits systematic exploration of model capabilities and constrains the development of reliable, domain-relevant applications. The challenges in materials science evaluation are particularly acute due to several factors. First, materials science encompasses a vast array of characterization techniques, each with its own specialized terminology, analytical principles, and interpretation methodologies~\cite{jiang_nlp_llm_matsci_2025}. Second, the field requires deep understanding of both theoretical concepts and practical experimental procedures, posing challenges in constructing evaluation items that accurately reflect real-world expertise~\cite{pei_lm_sustainability_2025}. Third, materials science often involves multimodal data analysis, including spectroscopic data, microscopic images, and structural information, requiring LLMs to integrate across diverse modalities effectively, a capability rarely captured in existing benchmarks~\cite{alampara_eval_ml_matsci_2025}.

To address this gap, this work proposes MatQnA, a benchmark dataset categorized according to material testing methodologies, aimed at evaluating the performance of mainstream LLMs in materials science tasks. To the best of our knowledge, this is the first multi-modal benchmark dataset specifically designed based on material characterization techniques. The key contributions of this paper are threefold:

\begin{itemize}
	\item \textbf{First}, we constructed a multi-category dataset materials focused on the field of materials science. The dataset is organized according to material characterization techniques, covering methods such as X-ray Photoelectron Spectroscopy (XPS), X-ray Diffraction (XRD), Scanning Electron Microscopy (SEM), and Transmission Electron Microscopy (TEM). It includes a large collection of domain-specific textual resources, such as journal articles and expert case studies corresponding to each testing method.
	\item \textbf{Second}, based on the collected materials, we employed OpenAI's GPT-4.1 API in combination with a human-in-the-loop validation process to construct a dataset that integrates model-assisted generation with manual verification. By leveraging preset prompt templates for question–answer pair generation and incorporating human oversight, we ensured both accuracy and reliability. This hybrid approach not only improves data quality but also enhances the scalability of the dataset, enabling broader coverage of interdisciplinary topics.
	\item \textbf{Third}, we conducted a preliminary evaluation of five mainstream multi-modal LLMs (GPT-4.1, Claude-sonnet-4, Gemini-2.5-flash, Qwen-2.5-VL-32B, Doubao Vision Pro 32K) across ten material characterization techniques. The results show that the most advanced models (e.g., GPT-4.1, Claude Sonnet 4, Gemini 2.5 Flash, and Doubao Vision Pro 32K) already achieve nearly 90\% accuracy on objective questions in materials data interpretation and analysis tasks, demonstrating strong application potential in this domain.
\end{itemize}

The remainder of this paper is organized as follows: Section 2 describes the dataset characteristics and question-answer pair types; Section 3 details the dataset construction workflow; Section 4 presents the evaluation results and analysis; Section 5 concludes with a discussion of implications and future directions; Section 6 provides a summary of the article, and Section 7 acknowledges the contributions of all participants.

\section{DATASET CHARACTERISTICS}

\subsection{Dataset Sources}
The dataset is primarily built upon materials science data accumulated from our proprietary platform, Scientific Compass. It encompasses a multi-source, heterogeneous corpus covering ten mainstream material characterization techniques. These techniques span the core dimensions of materials characterization, including structural analysis (XRD, XPS), microscopy (SEM, TEM, AFM), thermal analysis (DSC, TGA), spectroscopy (Raman, FTIR), and synchrotron analysis (XAFS).

\begin{figure}[!htbp]  
  \centering  
  \includegraphics[width=0.9\textwidth]{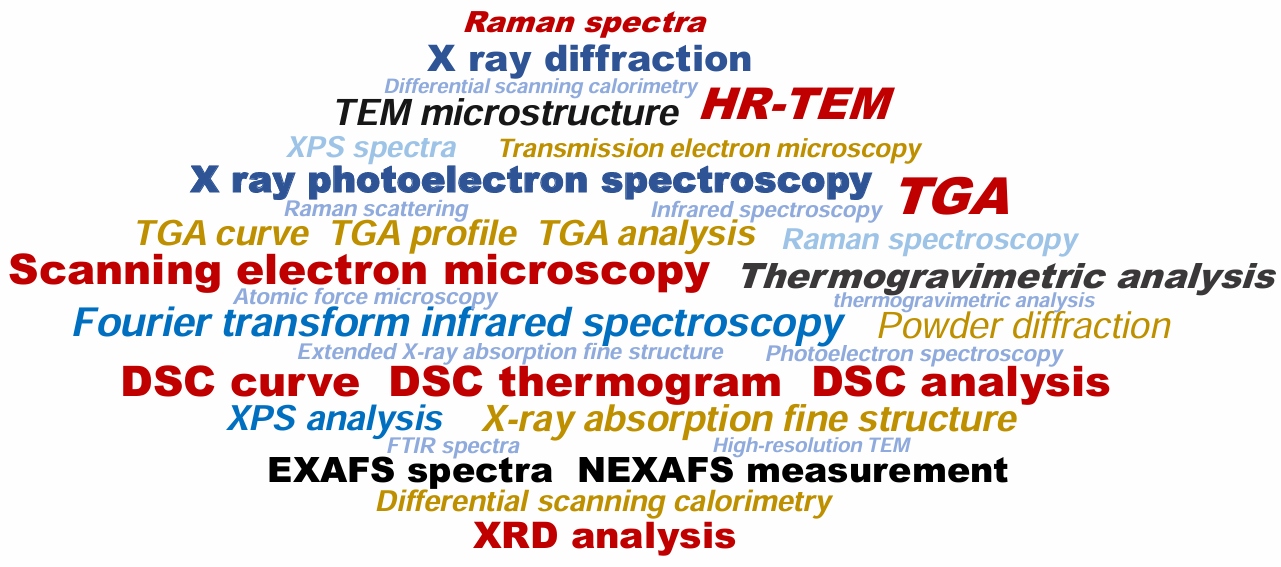}  
  \caption{Keywords for retrieving figures and texts related to materials characterization in the articles}  
  \label{fig:keyword}  
\end{figure}

In total, we collected data from more than 400 peer-reviewed journal articles published between late 2024 and early 2025. Materials characterization–related papers are selected through keyword matching, using terms such as "X-ray Photoelectron Spectroscopy," "X-ray Diffraction," "XPS," and "XRD," as illustrated in Figure \ref{fig:keyword}.

\begin{itemize}
	\item \textbf{Journal Articles}: We curated a selection of materials science papers published in high-impact domestic and international journals. The focus was on sections related to structural characterization, morphology analysis, spectral interpretation, and the correlation between figures and text. These data are academically rigorous and structurally standardized, providing a solid foundation for constructing expert-level question–answer pairs.
	\item \textbf{Expert Cases}: Our platform hosts a rich collection of analytical examples by experienced materials testing professionals. These cases cover spectrum interpretation, structural inference, parameter fitting, and selection of testing strategies. This subset reflects deeply structured domain knowledge and can be used to evaluate LLMs’ capabilities in complex reasoning and multi-step decision-making.
\end{itemize}

Overall, the dataset achieves a unified representation of modality diversity (images, spectra, text) and semantic richness (descriptive, inferential, and judgmental layers). It preserves the logical chain of materials science knowledge within experimental contexts, thereby providing a robust foundation for developing and evaluating large-scale models with domain-specific understanding and cross-modal reasoning abilities. The multimodal nature of the dataset is exemplified by the sample images (see Figure~\ref{fig:example_input_1} and Figure~\ref{fig:example_input_2}) and corresponding question-answer pair (see section \ref{sec:example_qa}), which demonstrate how text and image data are integrated to create comprehensive evaluation scenarios.

\subsection{Question-Answer Pair Types}
The selection of question types is critical for evaluating the capabilities of large language models (LLMs), as different formats impose distinct demands and can substantially affect the authenticity, fairness, and applicability of the results. After careful consideration, we adopted a hybrid format comprising multiple-choice questions (MCQs) and subjective questions (short-answer or open-ended), based on the following rationales:
\begin{itemize}
    \item \textbf{Balancing knowledge breadth and linguistic depth}: Subjective questions primarily assess the model’s ability to organize language and reconstruct factual information, while MCQs are more focused on evaluating the model’s recognition and judgment capabilities\cite{myrzakhan_open_eval_2024}\cite{grevisse2024shortanswer}.
    \item \textbf{Accuracy of fundamental knowledge}: Single-answer MCQs require models to identify the "best answer," enabling a more fine-grained analysis of their coverage of domain-specific knowledge\cite{rethinking_mcqa_2024}\cite{plosone2025mcqaccuracy}. 
    \item \textbf{Testing reasoning and expression}: Subjective questions allow us to evaluate whether the model can "explain clearly and correctly," providing insights into its logical organization and language articulation skills\cite{bernard_equator_2025}.
\end{itemize}


Compared with subjective questions, single-choice MCQs offer unique strengths that make them indispensable in model evaluation. Their objective and standardized scoring reduces subjectivity, ensuring fairness and reproducibility\cite{elsamanoudy2024meq}. Moreover, by constraining responses to one "best answer," MCQs allow precise measurement of domain knowledge while supporting large-scale benchmarking in an efficient and scalable manner.

\subsection{Dataset Category Distribution}
Table~\ref{tab:table1} provides an overview of the ten major material characterization techniques represented in the MatQnA dataset. The number of questions varies across techniques, with differing proportions of objective and subjective formats. All questions are primarily sourced from peer-reviewed journal articles, ensuring both reliability and domain relevance. Because each technique is methodologically independent—differing in principles, experimental procedures, and application contexts—answering the corresponding questions requires specialized expertise in materials science. By integrating diverse techniques and question types, MatQnA offers a rigorous and high-quality benchmark for assessing LLMs’ ability to address complex, interdisciplinary tasks in the field.

\begin{table}[htbp]
    \caption{Overview statistics of MatQnA dataset}
    \label{tab:table1} 
    \centering
    \begin{tabular}{cccccc}
        \toprule
        \multirow{2}{*}{Category} & \multirow{2}{*}{Total} & \multirow{2}{*}{Subjective} & \multirow{2}{*}{Objective} & \multicolumn{2}{c}{Data Source(\%)} \\
        \cmidrule{5-6}
        & & & & Journal Article & Expert Case \\
        \midrule
        AFM   & 266  & 148 & 118 & 93.2 & 6.8 \\
        DSC   & 273  & 146 & 127 & 90.1 & 9.9 \\
        FTIR  & 179  & 98  & 81  & 81.0 & 19.0 \\
        RAMAN & 265  & 146 & 119 & 84.9 & 15.1 \\
        SEM   & 811  & 441 & 370 & 91.9 & 8.1 \\
        TEM   & 755  & 420 & 335 & 95.8 & 4.2 \\
        TGA   & 324  & 181 & 143 & 96.6 & 3.4 \\
        XAFS  & 224  & 127 & 97  & 100.0 & 0.0 \\
        XPS   & 861  & 488 & 373 & 95.1  & 4.9 \\
        XRD   & 1010 & 554 & 456 & 95.0  & 5.0 \\
        \midrule
        TOTAL & 4968 & 2749 & 2219 & 93.5 & 6.5 \\
        \bottomrule
    \end{tabular}
\end{table}

\begin{itemize}
    \item \textbf{XPS (X-ray Photoelectron Spectroscopy) subset} evaluates a model’s ability to extract essential information on material composition, structural evolution, and binding energy characteristics from textual descriptions, while integrating reasoning based on critical spectral features such as peak positions, peak areas, and chemical states\cite{schultz2020practical}. The questions are divided into four categories: (1) chemical state identification and functional group recognition; (2) element identification and peak assignment; (3) peak fitting and structural interpretation; and (4) material type inference with structural discrimination.
    \item \textbf{XRD (X-ray Diffraction) subset} evaluates a model’s ability to accurately identify crystal structures, phase composition, and grain size from textual descriptions, while performing both qualitative and quantitative analyses based on diffraction peak positions, intensities, and full width at half maximum (FWHM) in XRD patterns. The questions are divided into four categories\cite{ali2022xrdminerals}: (1) qualitative phase analysis; (2) semi-quantitative phase analysis; (3) crystal structure parameter determination; and (4) grain size and stress analysis.
    \item \textbf{FTIR (Fourier Transform Infrared Spectroscopy) subset} evaluates a model’s ability to identify chemical bonds, functional groups, and molecular structural information from text, while analyzing characteristic peak wavenumber positions, intensities, and spectral assignments. The questions are divided into four categories\cite{ately2023atrftir}: (1) molecular structure characterization; (2) qualitative functional group identification; (3) chemical reaction process monitoring; and (4) material interface analysis.  

    \item \textbf{Raman (Raman Spectroscopy) subset} evaluates a model’s ability to interpret molecular vibration modes, structural disorder levels, and phase composition information, making integrated judgments based on Raman shifts, peak shapes, and relative intensities in spectra. The questions are divided into four categories\cite{liu2023ramanelectroceramics}\cite{zhang2014ramancarbon}: (1) material phase structure identification; (2) defect and stress state assessment; (3) carbon material structural type analysis; and (4) peak assignment with synthesis condition inference.  
    
    \item \textbf{DSC (Differential Scanning Calorimetry) subset} evaluates a model’s ability to identify thermal transition behaviors—such as glass transition, melting, crystallization, and reaction enthalpy—while analyzing parameters including endothermic/exothermic peak positions and enthalpy changes in DSC curves. The questions are divided into four categories\cite{schroeder2023analyticaldsc}\cite{he2018dscpolymers}: (1) thermal transition behavior identification; (2) peak position and enthalpy determination; (3) phase transformation process analysis; and (4) material stability with thermal property inference.  
    
    \item \textbf{TGA (Thermogravimetric Analysis) subset} evaluates a model’s ability to recognize mass change patterns during thermal decomposition, oxidation, and volatilization processes, while interpreting metrics such as weight-loss intervals, weight-loss rates, and residual mass in TGA curves. The questions are divided into four categories\cite{wang2022tgacarbon}\cite{muller2025tgareviewwood}: (1) mass-loss process identification and reaction stage determination; (2) thermal stability analysis; (3) residual mass and composition inference; and (4) decomposition mechanisms with reaction pathway interpretation.

   \item \textbf{SEM (Scanning Electron Microscopy) subset} evaluates a model’s ability to extract descriptive information about surface morphology, particle size distribution, and fracture characteristics, while analyzing structural features visible in images—such as dimensions, roughness, and morphology types. The questions are divided into five categories\cite{smith2022semquantitative}\cite{jones2023semmineral}: (1) particle size and distribution assessment; (2) surface structure and morphology identification; (3) defect or anomaly analysis; (4) microstructure–process correlations; and (5) integrated structural interpretation with material identification.  

    \item \textbf{TEM (Transmission Electron Microscopy) subset} evaluates a model’s ability to interpret crystal structures, defect types, and grain boundary behavior, while analyzing lattice fringes, diffraction spots, and atomic distributions in high-resolution images. The questions are divided into five categories\cite{williams2023temreview}: (1) crystal structure and defect analysis (dislocations, stacking faults, twinning, etc.); (2) nanoparticle morphology and size measurement; (3) phase boundary and interface structure studies; (4) diffraction pattern interpretation and structural identification; and (5) STEM imaging analysis (high-resolution imaging, Z-contrast).  

    \item \textbf{AFM (Atomic Force Microscopy) subset} evaluates a model’s ability to comprehend surface roughness, three-dimensional topography, and height variations, while analyzing topological characteristics, cross-sectional profiles, and surface morphology changes in AFM images. The questions are divided into five categories\cite{williams2023temreview}: (1) quantitative surface roughness analysis; (2) particle/grain size distribution measurement; (3) surface structure and morphological feature recognition; (4) effects of material processing on surface topography; and (5) phase distribution with interface identification.  

    \item \textbf{XAFS (X-ray Absorption Fine Structure) subset} evaluates a model’s ability to interpret local atomic environments, coordination numbers, and oxidation states, while analyzing absorption edge positions, pre-edge features, and extended fine structure oscillations\cite{newville2014fundamentals}. The questions are divided into four categories: (1) absorption edge analysis and chemical state determination; (2) coordination environment and local structure analysis; (3) extended fine structure interpretation and bond distance analysis; and (4) material composition and electronic structure inference.  
\end{itemize}

Owing to the methodological independence of experimental principles and application contexts across characterization techniques, the interpretation of related questions necessitates distinct and specialized knowledge structures in materials science. By systematically integrating diverse techniques and question formats, the dataset establishes a rigorous and high-value benchmark for assessing the capacity of large language models to address complex, domain-specific challenges and heterogeneous tasks within the field of materials characterization.

\section{DATASET CREATION}
\label{sec:others}

\begin{figure}[ht]
    \centering
    \includegraphics[width=1.0\textwidth]{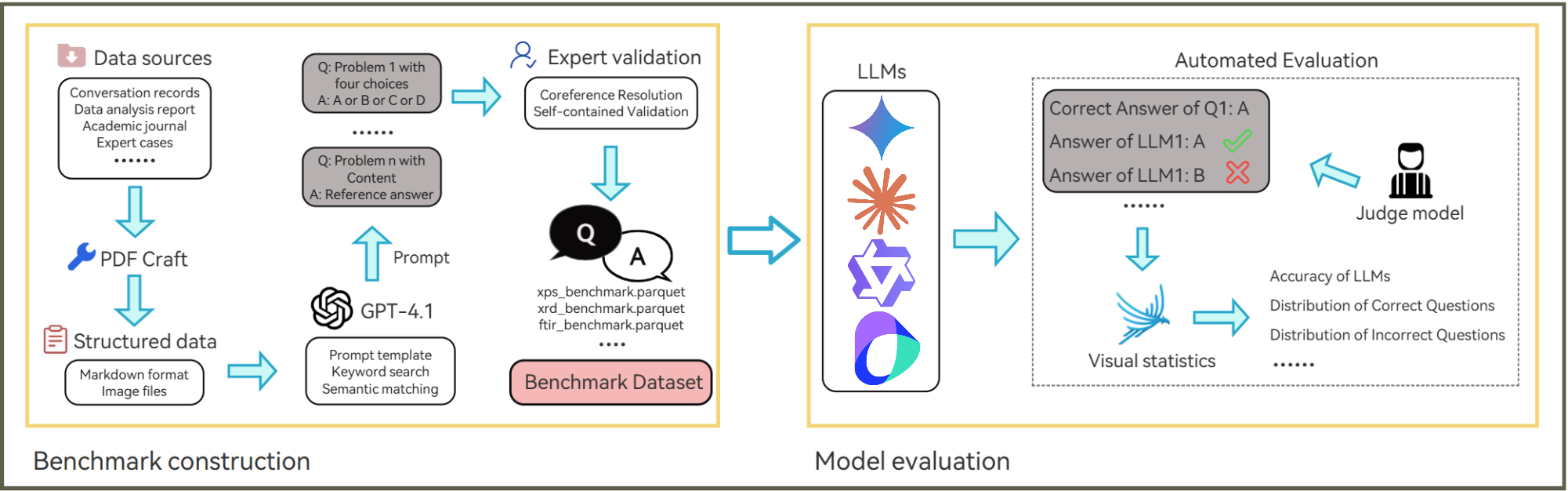}
    \caption{Generation of the MatQnA benchmark dataset and evaluation pipeline for Large Language Models}
    \label{fig:overview}
\end{figure}

We draw inspiration from the MatTools dataset~\cite{liu2025mattools} and adopt a hybrid approach that combines generation using OpenAI’s GPT-4.1 API with manual review and refinement. This involves the development of an automated dataflow based on LLMs, which will be detailed in this section. An overview of the process is illustrated in Figure~\ref{fig:overview}. The core ideas of this method are as follows:

\begin{itemize}
    \item Use LLMs to perform initial question–answer generation based on structured data sources, enabling efficient and large-scale production.
    \item Apply a lightweight human review and filtering mechanism to improve semantic quality, accuracy, and evaluation relevance.
    \item Ultimately construct a dataset of question–answer pairs that balances scale and quality, providing a more scientific and dynamic basis for evaluating the capability boundaries of large language models.
\end{itemize}

\subsection{Preprocessing}
\label{sec:preprocessing}
As shown in Table~\ref{tab:table1}, the dataset constructed in this study is primarily derived from journal articles archived on the Scientific Compass platform, which systematically categorizes papers according to material characterization techniques. For this work, the articles were organized into ten categories: XPS, XRD, SEM, TEM, AFM, DSC, TGA, FTIR, Raman, and XAFS. 

Since these documents were stored in PDF format, a structured preprocessing pipeline was necessary to enable automated question–answer (QA) extraction with large language models. Therefore, we employed the open-source tool PDF Craft, which leverages deep learning for multimodal PDF parsing. PDF Craft not only extracts text, images, and document structure, but also provides flexible output formats (Markdown, EPUB) and adaptive layout analysis, thereby ensuring more accurate and comprehensive content preprocessing.

\subsection{Benchmark Data Synthesis}
Given token and computational constraints, the first step in benchmark dataset construction was filtering irrelevant content. Using XPS as an example, we processed Markdown-formatted texts and associated image folders by building a keyword-based retrieval system and a text–image index aligned with document structure. Relevant fragments were then selected via semantic matching, with corresponding images retrieved through the index. These XPS-specific text–image pairs were fed into GPT-4.1, which, guided by predefined templates for multiple-choice and subjective questions (see Appendix or GitHub), automatically generated structured QA pairs. Each unit produced up to five questions, with fewer generated when content was limited, ensuring scientific relevance and quality.


Using this approach, we extracted QA pairs from preprocessed text and images for ten material characterization techniques (see Table~\ref{tab:table1}) and constructed the benchmark dataset in Parquet format to facilitate systematic evaluation. Detailed examples of QA pairs are presented in Section~\ref{sec:example_qa}, with representative instances based on input images shown in Figure~\ref{fig:example_input_1} and Figure~\ref{fig:example_input_2}, thereby illustrating both the quality and the complexity of the dataset.

\subsection{Post-Processing}
To ensure the scientific rigor and reliability of the benchmark dataset, we addressed issues arising from LLMs’ imperfect adherence to prompts and the generation of non-self-contained question–answer (QA) pairs. To this end, we implemented code-level constraints and validation mechanisms:

\begin{itemize}
    \item \textbf{Coreference Resolution}: Ambiguous references (e.g., “based on the given content,” “this figure,” “this text”) may lead to misinterpretation. We applied a regex-based normalization procedure to automatically detect and resolve such cases, thereby improving clarity and objectivity in evaluation items.
    \item \textbf{Self-Containment Enforcement}: Some QA pairs lacked sufficient image context, making the answers indeterminable from text alone. To address this, we introduced an image non-nullity check during data generation, ensuring that each item incorporates adequate multimodal context. This guarantees both interpretability and validity in downstream evaluation.
\end{itemize}

\subsection{Human Validation}
Human validation plays an essential role in ensuring the accuracy, domain relevance, and applicability of the generated dataset. We adopted a two-stage validation process to address limitations that may persist after automated post-processing. These include:
(1) over-specialization on specific figures or scenarios, limiting generalizability;
(2) distractor options misaligned with the core analytical intent;
(3) factual errors due to misused technical terms or semantic misunderstandings.

To address these issues, a team of materials science experts conducted a sampled review of the generated QA pairs. Each expert assessed the question stem, answer choices, model-generated explanation, and the associated context.The human validation stage serves two primary purposes:

\begin{itemize}
    \item \textbf{Ensuring Question Accuracy}: Experts verify terminological correctness, logical coherence in answer reasoning, and alignment with materials science principles.
    \item \textbf{Filtering for Relevance}: Questions with limited analytical value or weak domain relevance are removed, ensuring the final dataset is tightly focused on critical materials analysis scenarios.
\end{itemize}

\subsection{Example Analysis of Question-Answer Pairs}
\label{sec:example_qa} 

\begin{figure}[ht]
    \centering
    \includegraphics[width=0.65\textwidth]{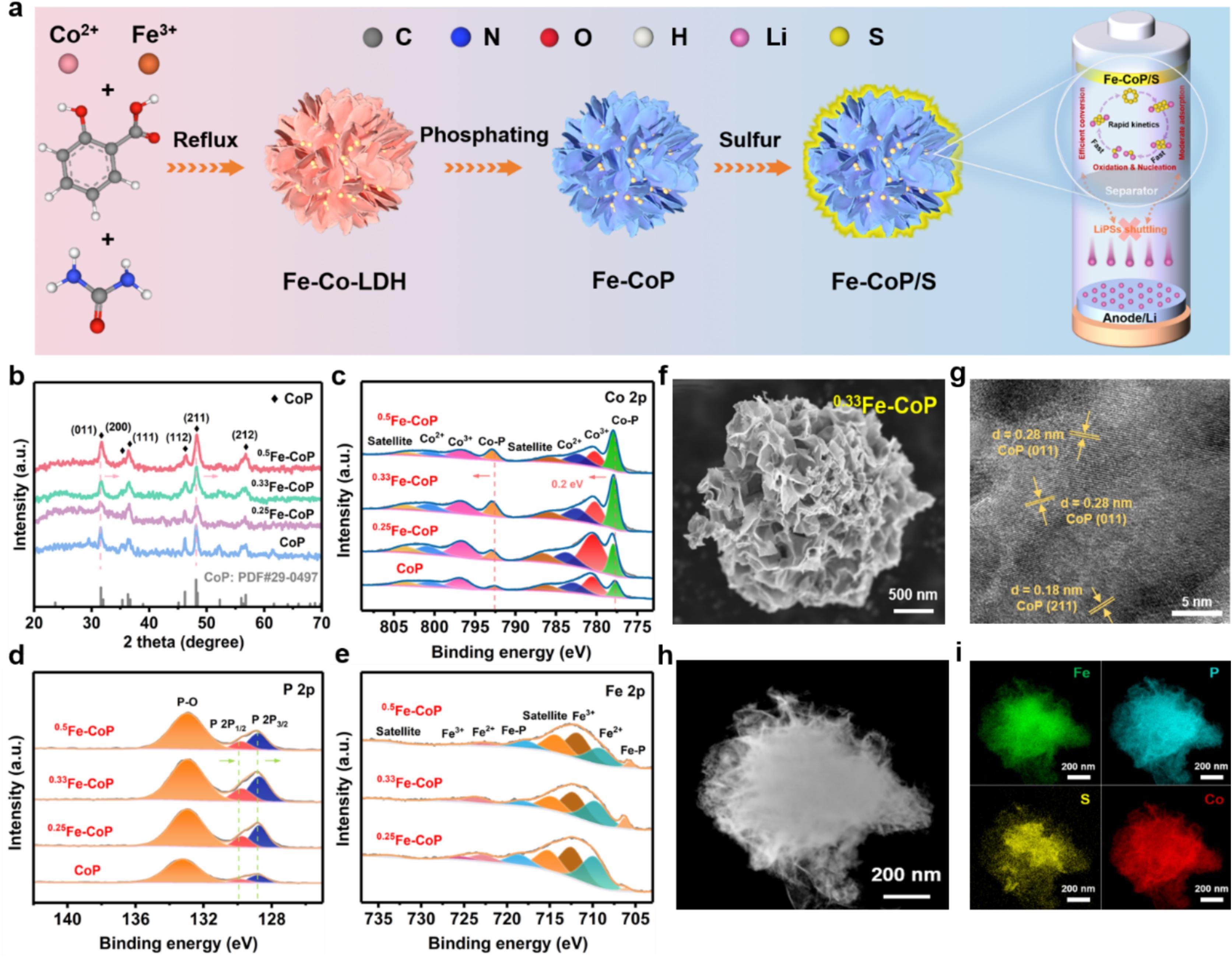}
    \caption{Example analysis image collected from  \cite{MU2025353}}
    \label{fig:example_input_1}
\end{figure}

\begin{figure}[ht]
    \centering
    \includegraphics[width=0.65\textwidth]{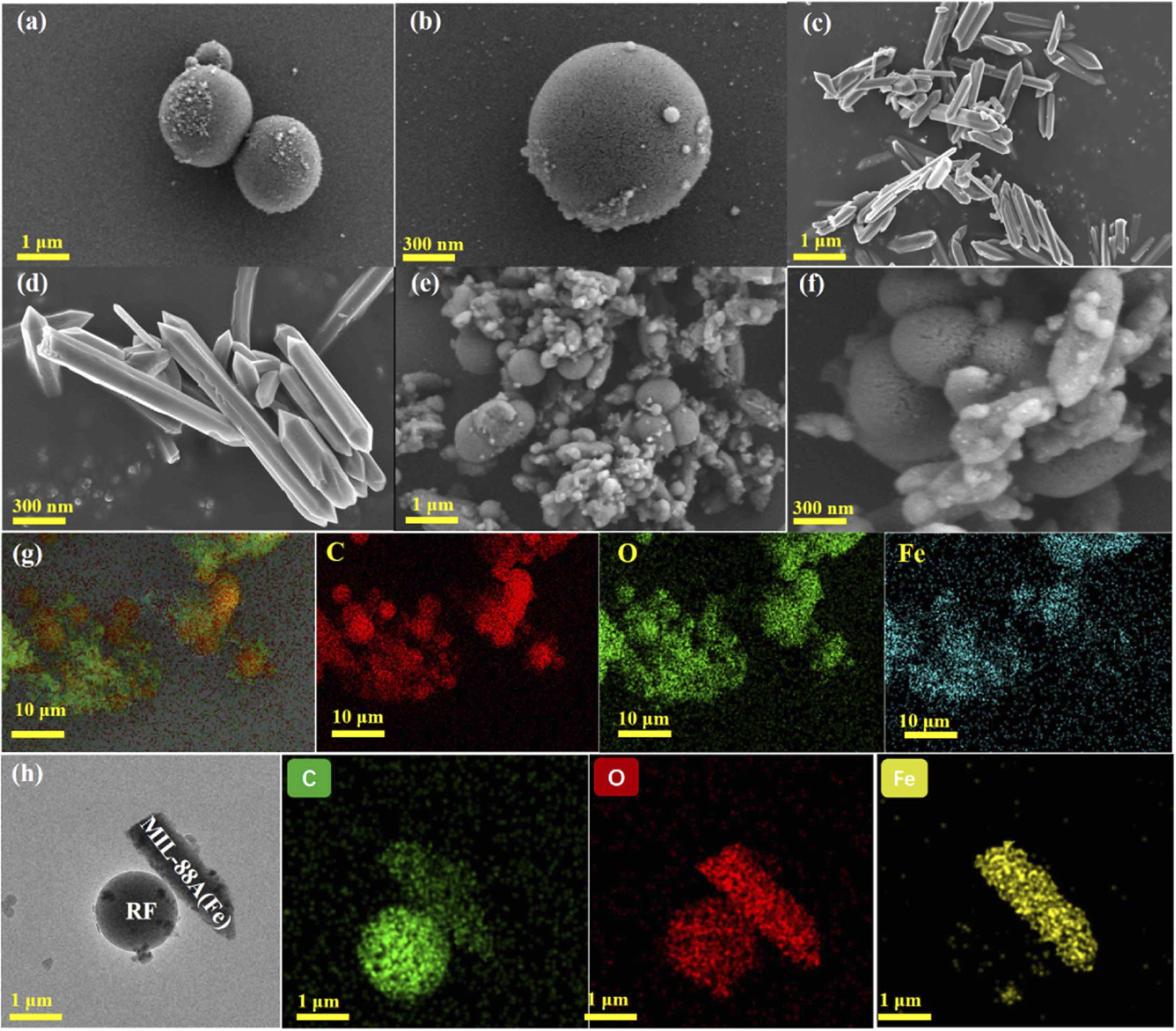}
    \caption{Example analysis image collected from \cite{DU2024124283}}
    \label{fig:example_input_2}
\end{figure}

Based on the provided images (Figure~\ref{fig:example_input_1} and Figure~\ref{fig:example_input_2}), we generated multiple question-answer pairs as follows:

\textbf{Question 1}: Analyzing the details of the main peak of $\text{CoP}$ in the diffraction spectrum: It is known that the main diffraction peaks of $\text{CoP}$ are located at 31.6$^{\circ}$, 35.3$^{\circ}$, 36.3$^{\circ}$, 46.2$^{\circ}$, 48.1$^{\circ}$, and 56.3$^{\circ}$, corresponding to the (011), (200), (111), (112), (211), and (212) crystal planes. The (011) and (211) peaks of a certain Fe-doped sample shift overall towards higher $2\theta$ angles compared to the pure $\text{CoP}$ sample, and no reflection peaks related to $\text{FeP}_x$ have appeared. Please determine which of the following analyses is the most reasonable?
   
\begin{itemize}[topsep=0pt, partopsep=0pt]
    \item[A.] The Fe element partially enters the $\text{CoP}$ lattice, forming a uniform solid solution of $\text{Fe-Co-P}$, and the uniform shift of all main peaks indicates an overall decrease in lattice constant.
    \item[B.] Fe precipitates in the form of separate $\text{FeP}$ or $\text{Fe}_2\text{P}$, forming a distinct second phase, leading to the splitting of the main peaks and the appearance of new peaks.
    \item[C.] Fe and Co are only mechanically mixed and do not enter the lattice, with the position of the main peaks remaining consistent with pure $\text{CoP}$ without significant shift.
    \item[D.] Fe only forms an epitaxial mixed crystal with $\text{CoP}$, resulting in random high and low angle splitting and irregular broadening of the main peaks.
\end{itemize}

\textbf{Question 2}: Focus on image content analysis, observe the XRD diffraction pattern, and analyze the $2\theta$ positions of all main peaks as well as their relative intensities and peak width distribution. Try to determine whether the material has a multiphase structure (such as CoP and $\text{Fe-CoP}$), and infer the effect of Fe element doping on the crystal structure and crystallinity based on the changes in peak positions and widths. Which of the following judgments is the most reasonable?

\begin{itemize}[topsep=0pt, partopsep=0pt]
    \item[A.] There is no significant movement of all main peak positions and no new peaks have appeared, indicating that there is no multiphase, and Fe doping has not changed the crystal structure.
   \item[B.] Some main peaks show slight low-angle shifts and peak broadening, with no new phase peaks appearing, indicating that Fe has been incorporated into the $\text{CoP}$ main phase lattice without phase separation.
    \item[C.] Multiple new peaks overlapping with the original peaks clearly appear, indicating that the material is a multiphase mixture of $\text{Fe-CoP}$ and $\text{CoP}$.
    \item[D.] All peaks shift to higher angles and the intensity of the main peaks increases, indicating that Fe doping leads to an increase in isotropic domain size and the generation of new metastable phases.
\end{itemize}

The AI models must demonstrate comprehensive understanding of image structure, accurately identify relevant sub-modules, and apply domain-specific knowledge to determine the correct answer.

\section{EXPERIMENTS AND RESULTS}
The models evaluated in this study include GPT-4.1, Claude 4, Gemini 2.5, Qwen 2.5, and Doubao. The performance metric used for evaluation is accuracy. The evaluation was conducted using the Phoenix platform, which leverages predefined environmental variables, model interfaces, and evaluation logic to facilitate the automated assessment of multiple large language models (LLMs) with a benchmark dataset. A detailed description of the evaluation process is provided in Figure \ref{fig:overview}.

\subsection{Overall Performance Analysis}

To ensure a fair and objective comparison across different models, our performance analysis focuses exclusively on objective questions (multiple-choice questions) from the dataset. While the dataset contains both objective and subjective question types, objective questions provide more standardized evaluation criteria and reduce potential bias from subjective scoring. Consequently, this approach enables direct comparison of model performance without the complexity introduced by evaluating open-ended responses.

\begin{table}[htbp]
	\caption{Evaluation results across material characterization techniques (Objective Questions Only)}
	\label{tab:results_no_prompts} 
    \centering
	\begin{tabular}{lccccc}
		\toprule
		Category & GPT-4.1 & Claude Sonnet 4 & Gemini 2.5 Flash & Qwen2.5 VL 72B & Doubao Vision Pro 32K \\
		\midrule
		AFM & 0.839 & 0.814 & 0.797 & 0.797 & 0.847 \\
		DSC & 0.866 & 0.890 & 0.858 & 0.811 & 0.898 \\
		FTIR & 0.951 & 0.926 & 0.938 & 0.926 & 0.951 \\
		RAMAN & 0.941 & 0.950 & 0.950 & 0.908 & 0.933 \\
		SEM & 0.897 & 0.903 & 0.892 & 0.868 & 0.900 \\
		TEM & 0.884 & 0.884 & 0.901 & 0.875 & 0.907 \\
		TGA & 0.902 & 0.916 & 0.902 & 0.902 & 0.909 \\
		XAFS & 0.907 & 0.907 & 0.907 & 0.876 & 0.907 \\
		XPS & 0.903 & 0.887 & 0.898 & 0.850 & 0.879 \\
		XRD & 0.906 & 0.908 & 0.904 & 0.853 & 0.884 \\
		\midrule
		\textbf{Overall} & \textbf{0.898} & \textbf{0.897} & \textbf{0.896} & \textbf{0.863} & \textbf{0.896} \\
		\bottomrule
	\end{tabular}
\end{table}

Based on this evaluation strategy, the results summarized in Table \ref{tab:results_no_prompts} indicate that all models exhibit strong capabilities in materials science analysis tasks, with overall accuracy scores ranging from 86.3\% to 89.8\%. Among them, GPT-4.1 achieves the highest overall performance (89.8\%), closely followed by Claude Sonnet 4 (89.7\%), while Gemini 2.5 Flash and Doubao Vision Pro 32K both reach 89.6\%. Qwen2.5 VL 72B, although slightly lower at 86.3\%, still demonstrates competitive performance in this domain

\begin{figure}[htbp]
    \centering
    \begin{minipage}{0.48\textwidth}
        \centering
        \includegraphics[width=\textwidth]{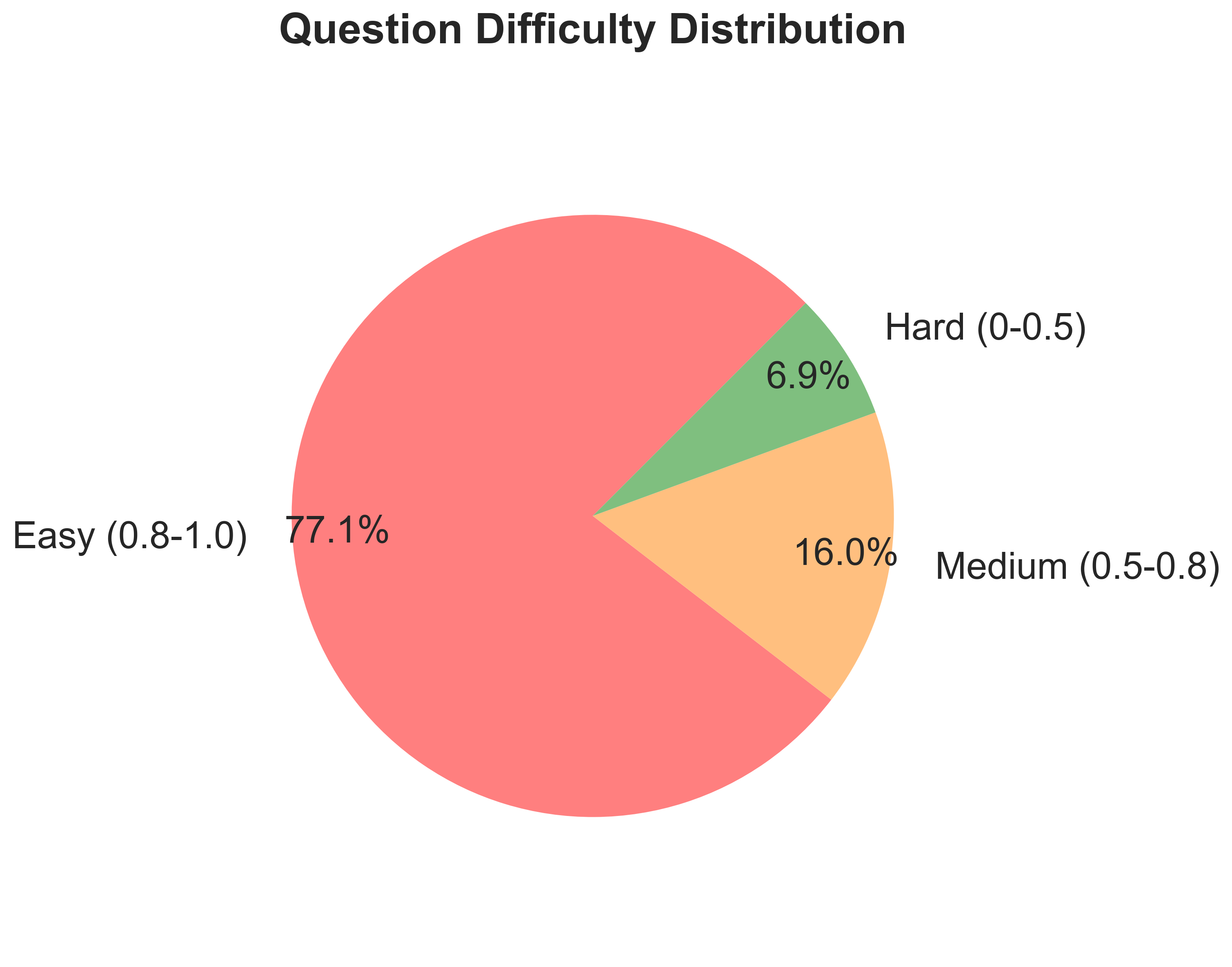}
        \caption{Overall difficulty distribution of objective questions, based on average accuracy scores across all evaluated models. Questions are categorized as Easy (accuracy $\geq0.80$), Medium (accuracy 0.50–0.79), or Hard (accuracy $<0.50$). The majority (77.1\%) are Easy, followed by 16.0\% Medium and 6.9\% Hard questions.}
        \label{fig:difficulty_distribution}
    \end{minipage}
    \hfill
    \begin{minipage}{0.48\textwidth}
        \centering
        \includegraphics[width=\textwidth]{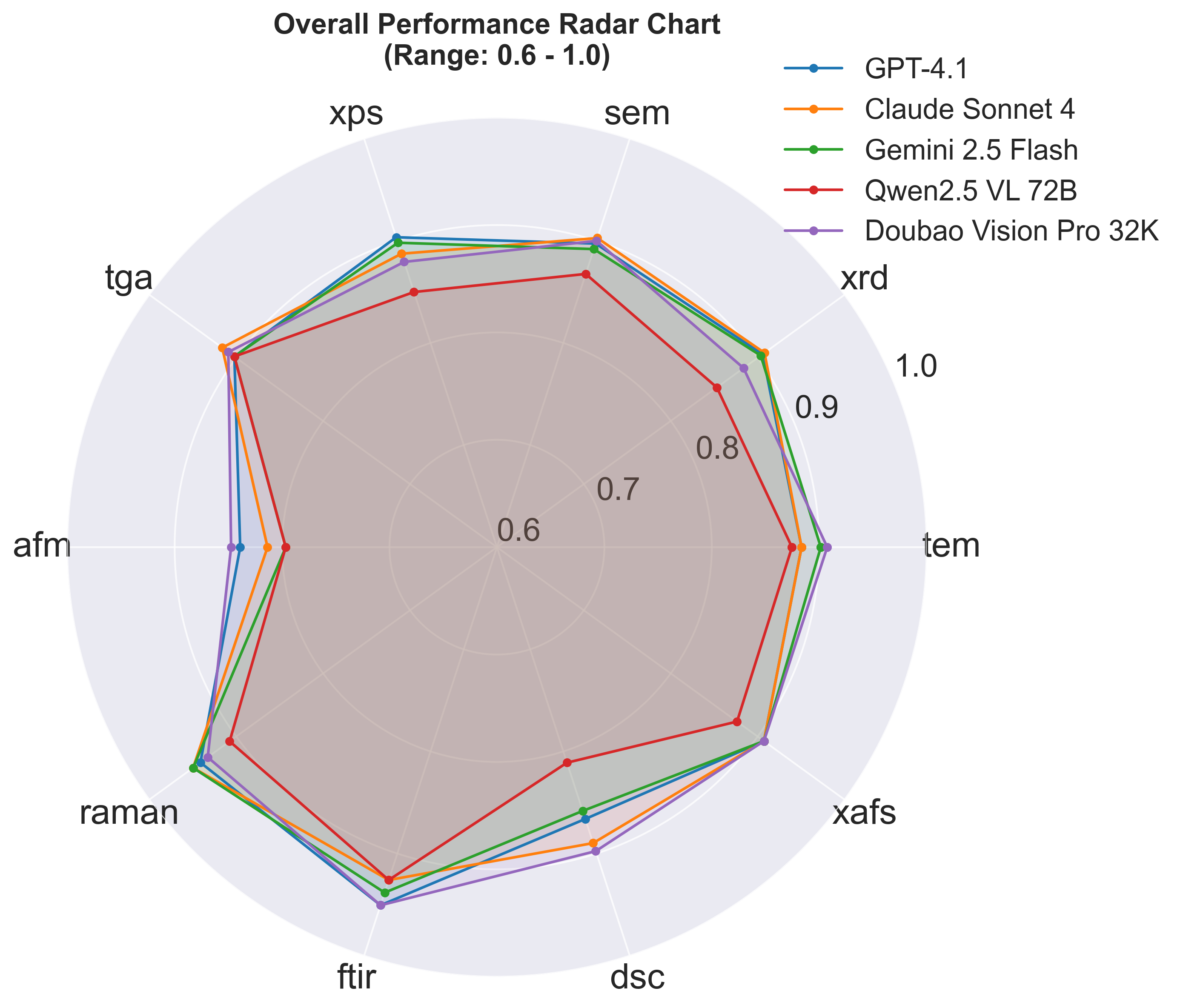}
        \caption{Overall performance radar chart of five LLMs across ten material characterization techniques (objective questions only). Each axis represents one technique, and values correspond to average accuracy scores (range: 0–1). Larger and more uniform polygons indicate stronger and more consistent performance. Detailed performance comparisons across models and techniques are discussed in the Results section.}
        \label{fig:radar_chart}
    \end{minipage}
\end{figure}

In addition, we statistically analyzed model performance across all material characterization categories and classified question difficulty based on average accuracy, as shown in Figure \ref{fig:difficulty_distribution}. Specifically, 77.1\% of the questions were categorized as easy (80–100\% accuracy), 16.0\% as medium difficulty (50–80\% accuracy), and 6.9\% as hard (0–50\% accuracy). While this distribution provides a solid foundation for evaluating current LLM capabilities, the relatively low proportion of hard questions (6.9\%) suggests a need for more challenging evaluation scenarios to push the boundaries of AI performance in materials science. The current benchmark effectively assesses basic knowledge and moderate reasoning capabilities, but may not fully capture the complexity of advanced analytical tasks encountered in cutting-edge materials research. Future iterations should incorporate more sophisticated problem types that require multi-step reasoning, cross-modal integration, and domain-specific expertise beyond what is currently represented in the dataset.

\subsection{Category-wise Model Performance Analysis}
Building upon these overall results, a more detailed analysis across different material characterization categories is presented in Figure \ref{fig:radar_chart}. GPT-4.1, Claude Sonnet 4, Doubao Vision Pro 32K, and Gemini 2.5 Flash achieved the highest performance, as indicated by large and uniform polygons across most techniques, particularly in FTIR, XAFS, TGA, and Raman. In contrast, Qwen2.5 VL 72B exhibited more variable performance with smaller, irregular polygons, showing lower accuracy in AFM, XAFS, and DSC categories. Detailed performance in each category is summarized as follows:
\begin{itemize}
    \item \textbf{High-Performance Categories}: FTIR, Raman, and TGA achieved high accuracy scores among the all evaluated models, exceeding 90\%. These results suggest that well-established characterization techniques with standardized protocols and clear interpretative frameworks are effectively handled by current LLMs. FTIR and Raman spectroscopy also demonstrated stronger performance (accuracy >92\%), indicating that spectroscopic analysis tasks involving pattern recognition and peak identification can be reliably processed by LLMs.
    \item \textbf{Challenging Categories}: AFM (Atomic Force Microscopy) was the most challenging task, with accuracy scores ranging from 79.7\% to 84.7\% across all models. This likely reflects the complex three-dimensional spatial reasoning and precise quantitative analysis required for AFM image interpretation. In the radar chart, AFM consistently appears as the innermost point of all model polygons, confirming its status as the most difficult characterization technique.
\end{itemize}

\subsection{Sub-category Model Performance Analysis}

\begin{figure}[htbp]
    \centering
    \includegraphics[width=1.0\textwidth]{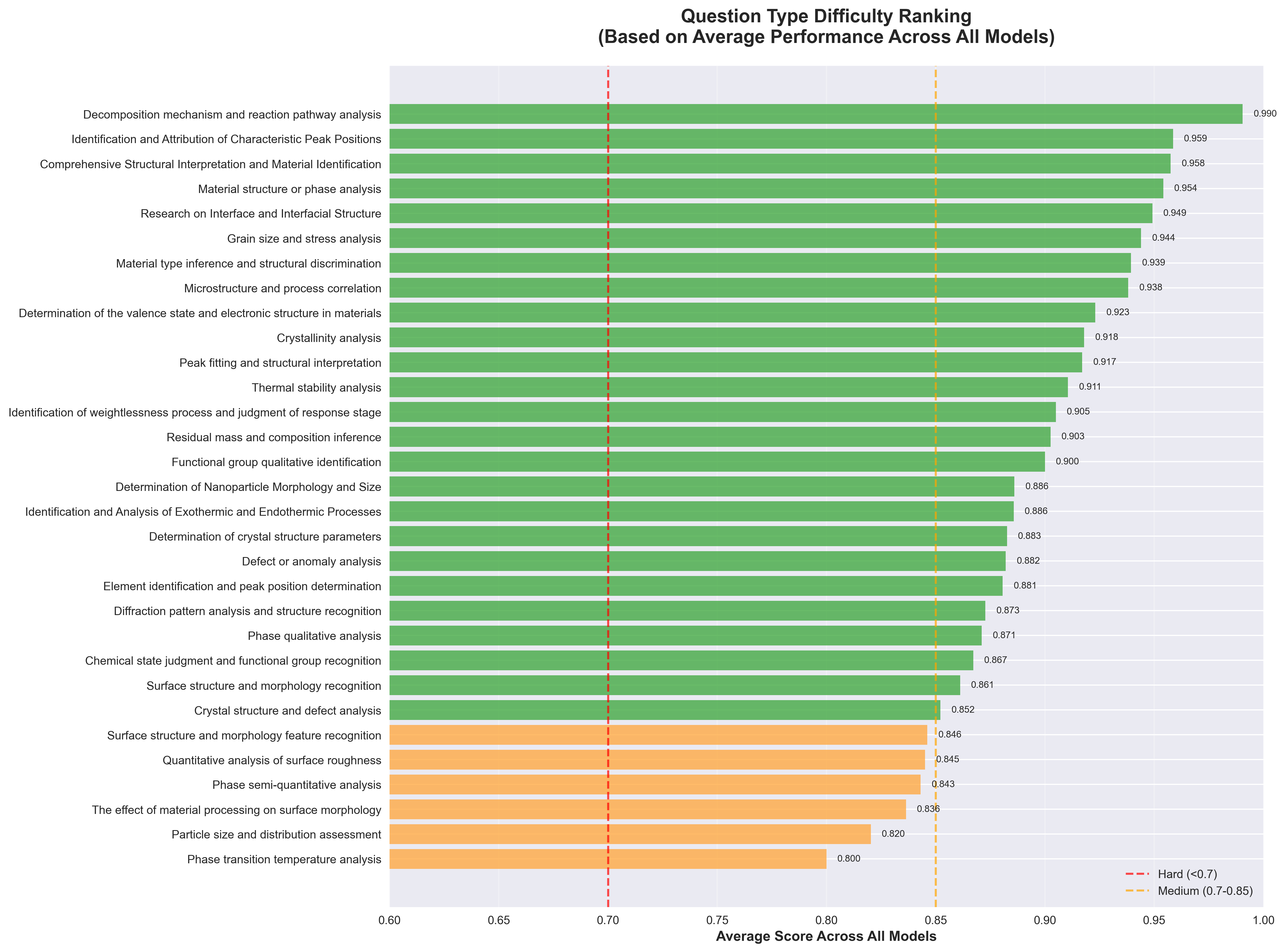}
    \caption{Question difficulty ranking across material analysis sub-categories based on average model accuracy. Higher scores indicate easier questions. Difficulty levels are defined as Hard ($<0.7$), Medium ($0.7$–$0.85$), and Easy ($>0.85$).}
    \label{fig:subcategory_difficulty}
\end{figure}

To provide deeper insights into performance patterns across material characterization sub-categories, we conducted a detailed analysis of question difficulty rankings based on average accuracy scores. Figure~\ref{fig:subcategory_difficulty} presents a horizontal bar chart showing the average accuracy scores of all evaluated models across different sub-categories. The results reveal several key observations regarding the difficulty distribution. 

\textbf{High-Performance Categories}\hspace{2mm} Most sub-categories exhibit excellent performance, with average accuracy scores above 0.85, suggesting that current LLMs achieve high accuracy on a wide range of materials analysis tasks. The top-performing sub-categories include:
\begin{itemize}
    \item \textbf{Decomposition mechanism and reaction analysis (0.990)}: This sub-category achieves the highest accuracy, which may be attributed to well-established theoretical frameworks and standardized analytical procedures for decomposition reactions.
    \item \textbf{Carbon identification and attribution (0.959)}: High accuracy likely reflects extensive training data for carbon materials and their well-characterized spectral signatures.
    \item \textbf{Comprehensive structural interpretation (0.958)}: Indicates that LLMs can effectively integrate multiple analytical perspectives and synthesize complex structural information.
    \item \textbf{Material structure or phase analysis (0.954)}: Strong performance suggests effective pattern recognition capabilities for crystalline structures.
    \item \textbf{Interface and interfacial studies (0.949)}: High accuracy in this sub-category indicates good understanding of surface and boundary phenomena.
\end{itemize}

\textbf{Medium-Difficulty Sub-categories}\hspace{2mm} Several sub-categories fall into the medium difficulty range (accuracy 0.70–0.85), representing areas where LLMs show room for improvement:

\begin{itemize}
    \item \textbf{Surface roughness quantitative analysis (0.846)}: Moderate performance likely reflects the complex three-dimensional spatial reasoning and precise quantitative requirements.
    \item \textbf{Phase semi-quantitative analysis (0.845)}: Challenges arise from the need to balance multiple analytical parameters and interpret relative intensities accurately.
    \item \textbf{Material processing effects analysis (0.843)}: This sub-category involves complex process–structure–property relationships, which may be challenging for current models.
    \item \textbf{Particle size distribution assessment (0.836)}: Statistical analysis of size distributions requires sophisticated reasoning capabilities, contributing to moderate performance.
    \item \textbf{Phase transition temperature analysis (0.820)}: The most challenging sub-category in the medium difficulty range, likely due to the need for precise temperature-dependent analysis and interpretation of thermal behavior.
\end{itemize}

\textbf{Performance Implications}\hspace{2mm} The difficulty ranking analysis provides insights for future model development and application optimization. High performance across most sub-categories indicates that current LLMs achieve high accuracy on routine materials analysis tasks, while the identified medium-difficulty areas highlight potential targets for improvement. The absence of sub-categories with average accuracy below 0.70 suggests that the benchmark effectively spans the practical difficulty range in materials science applications.

The analysis further reveals that spectroscopic techniques (FTIR, Raman) and structural analysis methods (XRD, XPS) generally achieve higher accuracy compared to microscopy-based techniques (AFM, SEM, TEM), which require more complex spatial reasoning and image interpretation. These results suggest that future model development should prioritize enhancing capabilities in quantitative analysis, spatial reasoning, and multi-parameter interpretation tasks.

\subsection{Fine-grained Model Performance Comparison}

\begin{figure}[htbp]
    \centering
    \includegraphics[width=0.8\textwidth]{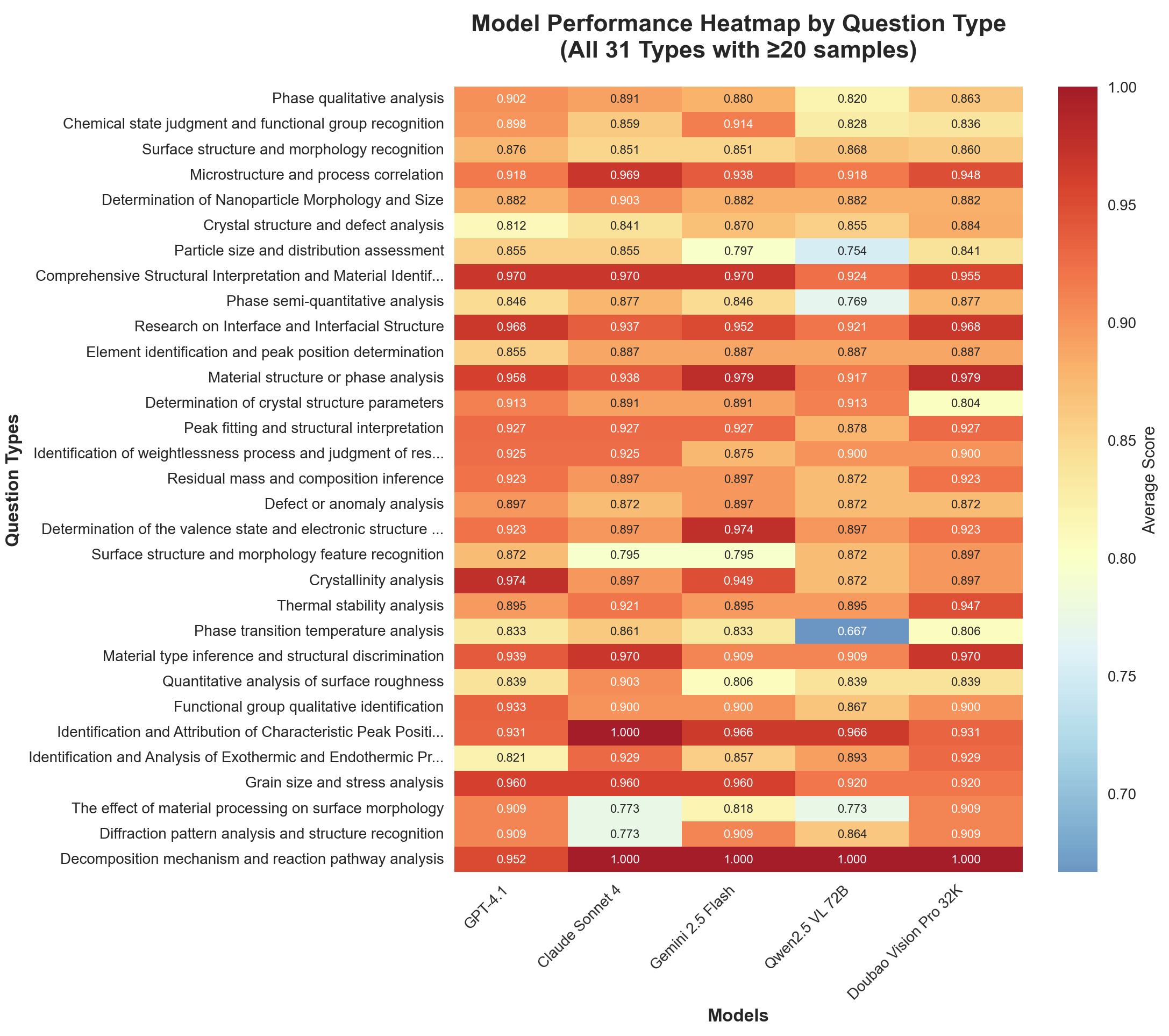}
    \caption{Heatmap of average performance scores across five AI models, grouped by 31 material characterization sub-categories ($\geq$20 samples per type). The color scale ranges from 0.0 (white) to 1.0 (dark red), indicating relative performance levels. This visualization highlights model-specific performance variations across sub-categories; detailed observations are discussed in the Results section.}
    \label{fig:model_subcategory_heatmap}
\end{figure}

To investigate model-specific strengths and weaknesses across material characterization sub-categories, we conducted a detailed performance comparison using a heatmap. Figure~\ref{fig:model_subcategory_heatmap} shows the average accuracy scores of five evaluated models across 31 distinct material characterization sub-categories, with each cell representing a model's performance on a specific sub-category. The heatmap reveals patterns of relative model performance across sub-categories.

\textbf{Overall Performance Hierarchy}\hspace{2mm} The analysis establishes a clear hierarchy among the evaluated models:

\begin{itemize}
    \item \textbf{Doubao Vision Pro 32K} achieved consistently high accuracy, with many cells showing scores above 0.95, as indicated by the dark red color scale. This model attained higher accuracy on tasks requiring multi-modal reasoning.
    \item \textbf{GPT-4.1} maintained strong performance across most sub-categories, with several accuracy scores exceeding 0.95, particularly in structural interpretation and phase analysis tasks.
    \item \textbf{Claude Sonnet 4 and Gemini 2.5 Flash} exhibited robust performance overall, with some categories reaching perfect scores (accuracy 1.000).
    \item \textbf{Qwen2.5 VL 72B} showed more variable performance, with lower scores in several challenging sub-categories as indicated by lighter-colored cells.
\end{itemize}

\textbf{Perfect Performance Categories}\hspace{2mm} Several sub-categories achieved outstanding accuracy across multiple models:

\begin{itemize}
    \item \textbf{Decomposition mechanism and reaction pathway analysis (accuracy 1.000)}: Four out of five models attained perfect scores, likely due to well-established theoretical frameworks and standardized analytical procedures.
    \item \textbf{Identification and Attribution of Characteristic Peak Positions (accuracy 1.000)}: Claude Sonnet 4 achieved perfect performance, reflecting strong pattern recognition capabilities in spectral analysis.
    \item \textbf{Comprehensive Structural Interpretation and Material Identification (accuracy 0.924–0.970)}: All models achieved consistently high scores, indicating robust capabilities in integrated structural analysis.
\end{itemize}

\textbf{Model-Specific Performance Patterns}\hspace{2mm} The analysis highlights distinct performance patterns for each model:

\begin{itemize}
    \item \textbf{Doubao Vision Pro 32K} achieved higher accuracy on multi-modal tasks requiring image interpretation and spatial reasoning, particularly in microscopy-based categories.
    \item \textbf{GPT-4.1} attained consistently high accuracy in structural analysis and pattern recognition tasks, especially in XRD- and XPS-related categories.
    \item \textbf{Claude Sonnet 4} performed particularly well in spectroscopic analysis and peak identification, achieving perfect scores in several spectral interpretation categories.
    \item \textbf{Gemini 2.5 Flash} showed balanced performance across most categories, with some notable peaks in reaction mechanism analysis.
    \item \textbf{Qwen2.5 VL 72B} performed adequately in basic structural analysis but had lower accuracy on complex quantitative and temperature-dependent tasks.
\end{itemize}

The results indicate that top-performing models, including GPT-4.1, Claude Sonnet 4, Gemini 2.5 Flash, and Doubao Vision Pro 32K, demonstrate consistently high and stable accuracy across diverse sub-categories, with several tasks even reaching perfect performance (accuracy = 1.000). In contrast, Qwen2.5 VL 72B shows more variable outcomes, with notable fluctuations and lower scores across multiple tasks, highlighting its comparatively less stable performance. Sub-categories such as decomposition mechanism and reaction pathway analysis, spectral peak identification, and comprehensive structural interpretation are particularly well-handled, reflecting the alignment between standardized analytical protocols and current LLM capabilities.

\section{LIMITATIONS AND FUTURE WORK}

While our evaluation provides valuable insights into LLM performance in materials characterization and analysis, several limitations may affect the interpretation and generalizability of the results.

\subsection{Dataset Quality Limitations}
Despite combining AI generation with expert validation, the dataset cannot guarantee complete accuracy. Some question–answer pairs may contain inconsistencies, reflecting the trade-off between scale and quality in domain-specific benchmarks. Moreover, most source papers are recent (late 2024–early 2025) and not publicly accessible, raising the possibility of data contamination if they overlap with model training corpora \cite{cheng_contamination_2025}. Such issues may bias evaluation outcomes and highlight the difficulty of assessing LLMs when training data sources are opaque \cite{sainz2023nlp}.

\subsection{Scope Limitations}
Our benchmark focuses on interpreting experimental data and figures, covering only a fraction of real-world materials science workflows, which also involve raw data parsing, database queries, literature synthesis, and multi-step reasoning. The question set is also skewed toward easier tasks, with only 6.9\% classified as hard, leaving advanced reasoning capabilities underexplored. Furthermore, due to computational constraints, we excluded specialized reasoning models that might perform better on complex analyses, limiting our understanding of the full range of LLM capabilities.

\subsection{Future Directions}
To address these limitations, future work should include contamination detection frameworks, dynamically updated or rewritten evaluation data, and contamination-prevention strategies\cite{deng2024_contamination}. It is crucial to design more challenging evaluation scenarios—featuring multi-step reasoning, cross-modal integration, and domain-specific complexity—and to develop AI agents that unify diverse analytical abilities. Additionally, incorporating real-world deployment considerations, such as inference speed, computational cost, and scalability, will better support the transition from prototype to production-ready AI systems in materials science.

\section{CONCLUSIONS}

In this paper, we introduce MatQnA, the first carefully-curated benchmark created to evaluate large language models in materials science applications. MatQnA is constructed from over 400 high-impact academic publications through a two-stage “LLM extraction – expert validation” pipeline, resulting in a benchmark resource of 4,968 questions, including 2,749 subjective and 2,219 objective items.As a preliminary step toward systematic evaluation, MatQnA aims to facilitate researchers and practitioners in assessing and improving LLM capabilities for materials characterization and analysis. Our findings highlight both the opportunities and challenges of applying LLMs in this domain, while emphasizing practical considerations such as data contamination, inference speed, computational costs, and scalability that must be addressed when transitioning from research prototypes to production-ready systems. Incorporating these real-world factors into future benchmark design will provide a more comprehensive evaluation framework, guiding both scientific progress and practical deployment.The dataset is publicly available at \url{https://huggingface.co/datasets/richardhzgg/matQnA}, and we encourage the research community to leverage this resource and contribute to its continuous improvement.

\section{ACKNOWLEDGMENTS}

We would like to express our sincere gratitude to all the laboratory engineers and analytical testing personnel who contributed to the construction of the MatQnA dataset. This work would not have been possible without their dedicated efforts in data collection, expert validation, and quality assurance. We particularly acknowledge the advices and contributions from the colleagues of Shiyanjia Lab (www.shiyanjia.com): Wendi Chen, Shijun Qiu, Weiyuan Ding, Xin Guo, Binbin Huang, Qiannan Ma, Ruiwen Niu, Jiawen Sun, Menghan Sun, Jiahao Wang, Meifang Wang, Tingmei Zhang.

\bibliographystyle{unsrtnat}
\bibliography{main}  






\end{document}